\documentclass[a4paper,fleqn]{cas-dc}

\usepackage[numbers]{natbib}

\usepackage{epstopdf}
\usepackage{tikz}
\usetikzlibrary{bayesnet}
\usepackage{algorithm}
\usepackage{algorithmic}
\usepackage{graphicx}
\usepackage{array}
\usepackage{hyperref}
\usepackage{cleveref}
\usepackage{caption}

\def\tsc#1{\csdef{#1}{\textsc{\lowercase{#1}}\xspace}}
\tsc{WGM}
\tsc{QE}
\tsc{EP}
\tsc{PMS}
\tsc{BEC}
\tsc{DE}

\begin{document}
\let\WriteBookmarks\relax
\def\floatpagepagefraction{1}
\def\textpagefraction{.001}
\shorttitle{Graph Convolutional Topic Model}
\shortauthors{Linh Ngo Van et~al.}

\title [mode = title]{A Graph Convolutional Topic Model for Short and Noisy Text Streams}                     



\author[1]{Ngo Van Linh}
\cormark[1]
\ead{linhnv@soict.hust.edu.vn}




\author[1]{Tran Xuan Bach}
\ead{tranxuanbach1412@gmail.com}
\author[1]{Khoat Than}
\ead{khoattq@soict.hust.edu.vn}
\address[1]{Hanoi University of Science and Technology, No. 1, Dai Co Viet road, Hanoi, Vietnam}


\begin{abstract}
Learning hidden topics from data streams has become absolutely necessary but posed challenging problems such as concept drift as well as short and noisy data. Using prior knowledge to enrich a topic model is one of potential solutions to cope with these challenges.  Prior knowledge that is derived from human knowledge (e.g. Wordnet) or a pre-trained model (e.g. Word2vec) is very valuable and useful to help topic models work better. However, in a streaming environment where data arrives continually and infinitely, existing studies are limited to exploiting these resources effectively. Especially, a knowledge graph, that contains meaningful word relations, is ignored. In this paper, to aim at exploiting a knowledge graph effectively, we propose a novel graph convolutional topic model (GCTM) which integrates graph convolutional networks (GCN) into a topic model and a learning method which learns the networks and the topic model simultaneously for data streams. In each minibatch, our method not only can exploit an external knowledge graph but also can balance the external and old knowledge to perform well on new data. We conduct extensive experiments to evaluate our method with both a human knowledge graph (Wordnet) and a graph built from pre-trained word embeddings (Word2vec). The experimental results show that our method achieves significantly better performances than state-of-the-art baselines in terms of probabilistic predictive measure and topic coherence. In particular, our method can work well when dealing with short texts as well as concept drift. The implementation of GCTM is available at \url{https://github.com/bachtranxuan/GCTM.git}.
\end{abstract}

\maketitle

\section{Introduction}

Topic modeling is a powerful approach to learn hidden topics/structures inside data. Latent Dirichlet allocation (LDA) \cite{blei2003latent} is one of the most popular topic models and has been used widely in a variety of applications such as text mining \cite{van2017effective}, recommender system \cite{le2018collaborative}, computer vision \cite{fei2005bayesian}, bioinformatics \cite{rogers2005latent}, etc. Recently, integrating external knowledge into LDA emerges as an effective approach to improve the origin. Prior knowledge, which is used in previous work, is derived from human knowledge (such as seed words \cite{fealda,kpasum}, Wordnet \cite{alkhodair2018improving}) or pre-trained models like word embeddings (Word2vec) \cite{zhao2017word,li2016topic} learnt from big datasets. Therefore, prior knowledge can enrich and improve the performances of topic models.

Meanwhile, developing an effective learning method for data streams has become absolutely necessary but posed challenging problems \cite{than2019make}. In this paper, we want to focus on two challenges. First, a learning method must adapt well to new data without revisiting past data. In order to solve this issue effectively, it must deal with the stability-plasticity dilemma \cite{mermillod2013stability,nguyen2018variational,kirkpatrick2017overcoming,ritter2018online,nguyen2019infinite}.  Particularly, in the streaming environment, data is big, arrives continually, and concept drift in which the statistics of data change dramatically can happen. A method should have a mechanism to keep acquired knowledge from learning on past data. This knowledge is useful to work on new data whose characteristics or patterns are similar to those from the past data. Simultaneously, it should be more plastic to learn a new concept that can appear any time. Second, noisy and sparse data that is prevailing in the streaming environment makes a lot of difficulties for learning methods \cite{mai2016enabling,ha2019eliminating,tuan2020bag}.  While sparse or short data does not provide a clear context, noisy data can mislead the methods. As a result, the generalization of learnt model can be limited. 

Exploiting a knowledge graph is one of the most potential solutions to cope with these challenges. It is obvious that a knowledge graph that comes from global human knowledge (e.g. Wordnet) or a pre-trained graph is valuable and useful to enrich a topic model to cope with short and noisy texts in the streaming environment. Moreover, a knowledge graph (such as Wordnet or a graph trained on a big dataset) contains meaningful word relations that seem to be static although concept drift can happen. Therefore, incorporating the graph into a topic model should be taken into consideration for data streams to deal with concept drift.

Although existing studies \cite{li2019integration,wang2019knowledge,yao2017incorporating,chen2013leveraging} can effectively exploit a knowledge graph in a static environment, they do not consider facing data streams and therefore do not work in the streaming environment where data arrives continually and infinitely. Meanwhile, several recent methods \cite{streamvb,masegosa17power,nguyen2018variational} can cope with data streams without revisiting past data. But they are limited to exploiting prior knowledge in general and a knowledge graph in particular. An implicit idea behind these methods is that a posterior distribution learnt in a minibatch is used as a prior for the following minibatch. As a result, in each minibatch, there are two prior distributions: The original prior distribution which is initialized in the first minibatch and the new prior which is derived from the posterior distribution learnt in the previous minibatch. Most of existing methods \cite{streamvb,nguyen2018variational,kirkpatrick2017overcoming,ritter2018online} only use the former in the first minibatch, then the latter replaces the former in the next minibatches.  A few methods \cite{masegosa17power,kpspakdd}  exploit them concurrently. However, they do not provide a way to exploit a knowledge graph.

There are two main issues that we want to address for an effective knowledge graph exploitation in the streaming environment. First, existing streaming methods ignore prior knowledge \cite{streamvb,kirkpatrick2017overcoming,nguyen2018variational} or require prior knowledge of a vector form \cite{kpspakdd,nguyen2020boosting}. In particular, they are unable to exploit prior knowledge of a graph form such as Wordnet or a pre-trained graph. For this problem, graph convolutional networks (GCN) \cite{kipf2017semi} can provide a potential solution to embed a knowledge graph in topic space. Thanks to which GCN can encode high-order neighbourhood relationship/structure, it can learn good graph embeddings to enrich topic models. Second, an automatic mechanism which controls the impact of a knowledge graph in each minibatch plays an important role in balancing the knowledge graph and old knowledge learnt from the previous minibatch. A suitable balancing mechanism can help exploit effectively both kinds of knowledge in practice and provide a potential solution to the stability-plasticity dilemma.

In this paper, we propose a novel model, namely \textit{Graph Convolutional Topic Model (GCTM)}, which integrates graph convolutional networks (GCN) \cite{kipf2017semi} into a topic model for data streams. We also develop a streaming method which simultaneously learns a probabilistic topic model and GCN in the streaming environment. GCTM has some benefits as follows:

\begin{itemize}
\item GCTM can effectively exploit a knowledge graph, which comes from human knowledge or a pre-trained model to enrich topic models for data streams, especially in case of sparse or noisy data.  We emphasize that our work first provides a way to model prior knowledge of graph form in the streaming environment. 
\item We also propose an automatic mechanism to balance the original prior knowledge and old knowledge learnt in the previous minibatch. This mechanism can automatically control the impact of the prior knowledge in each minibatch. When concept drift happens, it can automatically decrease the influence of the old knowledge but increase the influence of the prior knowledge to help our method deal well with the concept drift. 
\end{itemize}

We conduct experiments\footnote{ The implementation of GCTM is available at
https://github.com/bachtranxuan/GCTM.git.} to evaluate GCTM with both a human knowledge graph (Wordnet) and a graph built from pre-trained Word2vec. The extensive experiments show that our method can exploit the knowledge graph well to achieve better performances than the state-of-the-art baselines in terms of probabilistic predictive measure and topic coherence. In particular, our method outperforms significantly baselines when dealing with short texts as well as concept drift.

In the rest of the paper, the related work and background are briefly summarized in section 2. Section 3 presents our proposed model and method along with some discussions about them.  We conduct experiments and analyse experimental results in section 4. The conclusion is drawn in section 5.

\section{Related Work and Background}
In this section, we review streaming learning methods and graph convolutional networks, then present how some streaming methods apply to LDA.

\subsection{Related Work}

Recently, learning from data streams has been studied intensively and several methods have been proposed to solve characteristic problems in streaming environments such as avoiding revisiting all past data \cite{lpp,streamvb,populationdis}, adapting to concept drift \cite{masegosa17power}, reducing catastrophic forgetting \cite{kirkpatrick2017overcoming,nguyen2018variational}, etc. They have achieved some good results in both practice and theory \cite{cherief2019generalization}.

With regard to learning manner, existing studies can be divided into two major directions: Stochastic optimization problem and recursive Bayesian learning. The first direction \cite{lpp,populationdis,khan2018fast} uses stochastic natural Gradient ascent to maximize the expectation of the likelihood. Stochastic variational inference (SVI) \cite{lpp} optimizes an empirical expectation on the whole dataset and therefore requires the existence of a full dataset with a fixed number of data instances. This assumption is unsuitable for streaming environments where the data can arrive infinitely. Population variational Bayes (PVB) \cite{populationdis} alleviates this problem by another assumption. It assumes that the data is generated from a population distribution and we can sample a fixed number (the size of the population) $S$ of data instances at a time for computing and optimizing the expectation. However, $S$ must be tuned manually to achieve good performance. In the other direction, the recursive Bayesian approach \cite{streamvb,masegosa17power,kpspakdd,kirkpatrick2017overcoming,nguyen2018variational} bases on an implicit idea that a posterior distribution learnt in the previous minibatch is used to form a new prior distribution in the current minibatch. Several methods such as Streaming variational Bayes (SVB) \cite{streamvb}, Hierarchical power prior (HPP) \cite{masegosa17power}, Variational continual learning (VCL) \cite{nguyen2018variational} use the full Bayesian approach to approximate the posterior distribution, while Elastic weight consolidation (EWC) \cite{kirkpatrick2017overcoming} and its variants \cite{AljundiBERT18,ritter2018online} base on the maximum a posterior (MAP) estimate. Many methods \cite{ZenkePG17,kirkpatrick2017overcoming,nguyen2018variational,ritter2018online} in this direction are proposed to make neural networks deal with the changes of tasks over time in streaming environments. In our work, we only consider methods that work well on topic models without changing task.

Meanwhile, to mitigate the problems of noisy and short texts, there are three main approaches: Exploiting external knowledge, aggregating short texts, and developing new suitable models for short texts. The first approach \cite{qiang2017topic,zhao2017word,li2017enhancing} uses word embedding to enrich information and therefore achieves significant improvements in comparison with the original models. However, existing studies in this approach have not considered developing a method for data streams. They merely focus on a static environment without changing data. Moreover, a knowledge graph is also ignored.  In the second approach, several methods \cite{mehrotra2013improving,quan2015short,bicalho2017general,mai2016enabling} modify the document input of conventional topic models to enhance word co-occurrence information. A strategy of aggregating short texts to a longer text is widely used in practice. The third approach \cite{cheng2014btm,yang2018topic,xu2018topic,yao2019graph,tuan2020bag} aims to propose a new model which is more suitable to model word-occurrence information for short texts instead of utilizing conventional topic models. However, both the second and third approaches ignore external knowledge in the streaming environment. In our work, we focus on developing an effective method to exploit a knowledge graph for data streams. We emphasize that our method can apply to not only LDA but also a wide range of existing topic models. It means that our method can improve performances of existing models in both the second and third approaches. 

In terms of exploiting prior knowledge in the streaming environment, KPS (Keeping prior in streaming Bayesian learning) \cite{kpspakdd,nguyen2020boosting} takes external knowledge into consideration, while the remaining methods neglect it. In the standard view of Bayesian approach, a prior distribution does not play an important role when data is big enough. It seems to be the main reason why almost existing methods ignore prior knowledge in streaming environments. Although KPS shows a vital role of prior knowledge for data streams, it remains two main drawbacks: The limit of prior knowledge form and a lack of balancing mechanism between prior knowledge and old knowledge learnt from previous data. Recently, our other work \cite{bach2020dynamic} aims to exploit external knowledge of different forms (such as vector, matrix) for data streams. However, it lacks an effective solution to capture relation between nodes in a knowledge graph. 

Recently, graph convolutional networks (GCN) \cite{kipf2017semi} emerges as an effective and efficient solution to learn graph embeddings. In practice, many previous studies show that GCN can work well in a wide variety of applications such as node classification \cite{kipf2017semi}, text classification \cite{yao2019graph}, machine translation \cite{bastings2017graph}, etc. In a recent work \cite{zhu2018graphbtm}, GCN is used in an inference network to learn a representation of a word co-occurrence graph for inferring local variables (the topic proportion of a biterm subset) better in the biterm topic model. However, this work does not consider using prior knowledge to enrich a topic model. Our work aims at a different goal. We exploit a knowledge graph to infer directly global variables (topics) instead of local variables. We emphasize that our work provides a general solution with a knowledge graph to improve existing models.

\subsection{Overview of Streaming Learning Methods for LDA \label{LDA}}
In this subsection, we briefly present LDA and learning methods that help LDA work in the streaming environment. 

Suppose that a document $d$ in a dataset contains $N_d$ words. A topic is defined by a distribution over $V$ words of the vocabulary.  LDA models $K$ hidden topics in the dataset and topic proportion of each document.  Let $\beta_1,...,\beta_K$ be $K$ hidden topics, $\theta_d$ be topic proportion of document $d$, and $z_{dn}$ be topic assignment of word $n$ in document $d$. LDA uses two Dirichlet distributions with hyerparameters $\eta$ and $\alpha$ to generate topics and topic proportions respectively. Both $\alpha$ and $\eta$ are often selected manually. The graphical representation of LDA is shown in Figure \ref{fig:graph_lda_model}. The generative process of LDA is as follows:
 
\begin{figure}
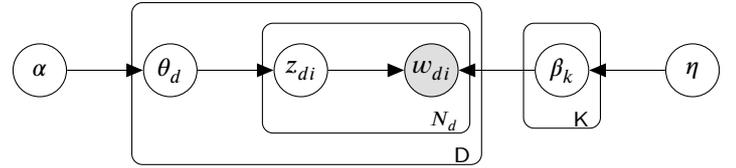

\centering
      \tikz{ %
        \node[latent] (alpha) {$\alpha$} ; %
        \node[latent, right=of alpha] (theta) {$\theta_d$} ; %
        \node[latent, right=of theta] (z) {$z_{di}$} ; %
        \node[obs, right=of z] (w) {$w_{di}$} ; %
        \node[latent, right=of w] (beta) {$\beta_k$} ; %
        \node[latent,right=of beta] (eta) {$\eta$};%
        \plate[inner sep=0.14cm, xshift=0cm, yshift=0.12cm] {plate1} {(z) (w)} {$N_d$}; %
        \plate[inner sep=0.14cm, xshift=0cm, yshift=0.12cm] {plate2} {(theta) (plate1)} {D}; %
        \plate[inner sep=0.14cm, xshift=0cm, yshift=0.12cm] {plate3} {(beta)} {K};
        \edge {alpha} {theta} ; %
        \edge {theta} {z} ; %
        \edge {z,beta} {w} ; %
        \edge {eta} {beta};
      }
	\caption{The graphical representation of Latent Dirichlet Allocation (LDA)}
	\label{fig:graph_lda_model}
\end{figure}

\begin{enumerate}
\item Draw topics $ \beta_{k} \sim \text{Dirichlet}(\eta) $ for $ k \in [1, K] $
\item For each document $ d  $:
	\begin{enumerate}
		\item Draw topic proportions $ \theta_{d} \sim \text{Dirichlet}(\alpha) $
		\item For each word $w_{dn}$: 
		\begin{enumerate}
			\item Draw topic assignment $z_{dn} \sim \text{Multinomial}(\theta_d) $
			\item Draw word $w_{dn} \sim \text{Multinomial}(\beta_{z_{dn}})$
		\end{enumerate}
	\end{enumerate}
\end{enumerate}

Training LDA is often divided into two phases: Inferring local variables ($z_d$ and $\theta_d$) for each document $d$ and learning global variable ($\beta$) shared among all documents. Almost streaming learning methods for LDA are the same in the former but are different in the latter. SVB \cite{streamvb}, PVB \cite{populationdis}, and HPP \cite{populationdis} approximate the posterior distribution of $\beta$ by a variational distribution $q(\beta|\lambda)$ in full Bayesian manner. Note that VCL and SVB are the same \cite{TheisH15,nguyen2018variational,cherief2019generalization} when they are applied to a conjugate model like LDA. Moreover, VCL \cite{nguyen2018variational} focuses on the problem of task changing, therefore, we do not consider in this paper.  We will briefly present the learning algorithms of SVB, PVB and SVB-PP (a simple version of HPP) for LDA. 

Suppose that in the streaming environment, the documents arrive continually and are collected in subsets (minibatches) with $D$ documents. For each minibatch $t$, mean-field variational inference is used to approximate the true posterior distributions of variables by variational distributions:
\begin{align}
q(\beta,\theta_d, z_d) =  \prod_{k=1}^K q(\beta_k|\lambda_k) \prod_{d=1}^D \left( q(\theta_d|\gamma_d) \prod_{n=1}^{N_d} q(z_{dn}|\phi_{dn}) \right)
\end{align} 
where:  $q(\beta_k|\lambda_k) = Dirichlet(\lambda_k)$, $q(\theta_d|\gamma_d) = Dirichlet(\gamma_d)$ and $q(z_{dn}|\phi_{dn}) = Multinomial (\phi_{dn})$ ($\lambda$, $\gamma$, and $\phi$ are variational parameters). Let $n_{dv}$ be the frequency of words $v$ in document $d$. The learning process of SVB, SVB-PP, and PVB are presented in Algorithms \ref{algo:SVB}, \ref{algo:SVB-PP}, and \ref{algo:PVB} respectively, where $E_q[\log \theta_{dk}] = \psi(\gamma_{dk})-\psi (\sum_{k=1}^K (\gamma_{dk}))$ and $E_q[\log \beta_{kv}] = \psi(\lambda_{kv})-\psi (\sum_{v=1}^V (\lambda_{kv}))$ ($\psi$ is a digamma function). The three methods have the same algorithm (Algorithm \ref{algo:LocalVB}) for doing inference local variables.  

\begin{table}[pos=!ht]
\begin{minipage}{0.45\textwidth}
\begin{algorithm}[H]
    \caption{LocalVB(d,$ \lambda $)}
	\label{algo:LocalVB}
\begin{algorithmic}
\STATE {Initialize: $ \gamma_d $}
\WHILE {$ (\gamma_d, \phi_d) $ not converged}
\STATE {$ \forall (k,v) $ set $ \phi_{dkv} \varpropto exp(E_q[\log \theta_{dk}] + E_q[\log \beta_{kv}]) $ (normalized across k)}
\STATE{ $ \forall k $, $ \gamma_{dk} \leftarrow \alpha_k + \sum_{v=1}^V \phi_{dkv} n_{dv} $}
\ENDWHILE
\RETURN {$ \gamma_d, \phi_d $}
\end{algorithmic}
\end{algorithm} 
\end{minipage} 
\begin{minipage}{0.45\textwidth}
\begin{algorithm}[H]
    \caption{SVB}
	\label{algo:SVB}
\begin{algorithmic}
\REQUIRE{Hyper-parameter $\alpha, \eta$}
\ENSURE{A sequence $ \lambda^{(1)}, \lambda^{(2)}, \ldots $}
\STATE {Initialize: $ \forall (k,v), \lambda_{kv}^{(0)} \leftarrow \eta_{kv} $}
\FOR  {$t = 0, 1,\ldots$} 
\STATE {Collect new data minibatch $D$}
\FOR {\textbf{each} document $d$ in $D$}
\STATE {$ (\gamma_d, \phi_d) \leftarrow LocalVB(d,\lambda) $}
\ENDFOR
\STATE {$ \forall (k,v), \lambda_{kv}^{t} \leftarrow \lambda_{kv}^{t-1} + \sum_{d\, in\, C} \phi_{dkv} n_{dv} $}
\ENDFOR
\end{algorithmic}
\end{algorithm}
\end{minipage}
 \\
\begin{minipage}{0.45\textwidth}
\begin{algorithm}[H]
    \caption{SVB-PP}
	\label{algo:SVB-PP}
\begin{algorithmic}
\REQUIRE{Hyper-parameter $\alpha, \eta, \rho_t$}
\ENSURE{A sequence $ \lambda^{(1)}, \lambda^{(2)}, \ldots $}
\STATE {Initialize: $ \forall (k,v), \lambda_{kv}^{(0)} \leftarrow \eta_{kv} $}
\FOR  {$t = 0, 1,\ldots$} 
\STATE {Collect new data minibatch $D$}
\FOR {\textbf{each} document $d$ in $D$}
\STATE {$ (\gamma_d, \phi_d) \leftarrow LocalVB(d,\lambda) $}
\ENDFOR
\STATE {Compute: $ \tilde{\lambda} = \rho_t \lambda_{kv}^{t-1}+ (1-\rho_t)\eta_{kv} $}
\STATE {$ \forall (k,v), \lambda_{kv}^{t} \leftarrow \tilde{\lambda} + \sum_{d\, in\, C} \phi_{dkv} n_{dv} $}
\ENDFOR
\end{algorithmic}
\end{algorithm} 
\end{minipage} \hspace{10pt}
\begin{minipage}{0.45\textwidth}
\begin{algorithm}[H]
    \caption{PVB}
	\label{algo:PVB}
\begin{algorithmic}
\REQUIRE{Hyper-parameter $\alpha, \eta, \rho_t, \tau_0, \kappa, B$}
\ENSURE{A sequence $ \lambda^{(1)}, \lambda^{(2)}, \ldots $}
\STATE {Initialize: $ \forall (k,v), \lambda_{kv}^{(0)} \leftarrow \eta_{kv} $}
\FOR  {$t = 0, 1,\ldots$} 
\STATE {Collect new data minibatch $D$}
\FOR {\textbf{each} document $d$ in $D$}
\STATE {$ (\gamma_d, \phi_d) \leftarrow LocalVB(d,\lambda) $}
\ENDFOR
\STATE {Compute: $ \rho_t ={(\tau_0+t)}^{-\kappa} $}
\STATE {Compute: $ \tilde{\lambda} = \eta_{kv}+ \frac{\alpha}{B}  \sum_{d\, in\, C} \phi_{dkv} n_{dv} $}
\STATE {$ \forall (k,v), \lambda_{kv}^{t} \leftarrow \rho_t\tilde{\lambda}_{kv} + (1-\rho_t)\lambda_{kv}^{t-1} $}
\ENDFOR
\end{algorithmic}
\end{algorithm}
\end{minipage}
\end{table}


\section{Graph Convolutional Topic Model (GCTM) for Data Streams}

In this section, we first present a our proposed model, then develop a learning method that learns our model from the streaming environment. Finally, we discuss some advantages of our model.

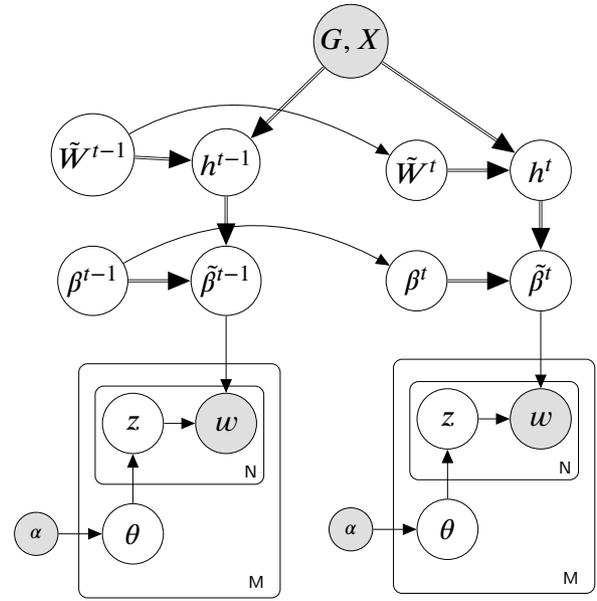
\begin{figure}[pos=!ht] 
\centering
\resizebox{0.45\textwidth}{!}{
\begin{tikzpicture}
\node[latent,scale=1.4](b){$\tilde{\beta}^{t-1}$};%
\node[latent, left=1 of b,scale=1.4](p){$\beta^{t-1}$};%
\node[latent, above=0.8 of p,scale=1.4](wg){$\tilde{W}^{t-1}$};%
\node[latent, above=0.8 of b,scale=1.4](g){$h^{t-1}$};%
\node[obs, below=1.25 of b,scale=1.4](w){$w$};%
\node[latent, left=0.5 of w,scale=1.4](z){$z$};%
\node[latent, below=0.8 of z,scale=1.4](theta){$\theta$};%

\node[obs, left=0.7 of theta](a){$\alpha$};%
\plate[inner sep=0.1cm,yshift=0cm]{document}{(z) (w)}{N};%
\plate[inner sep = 0.25cm,yshift=0.1cm]{corpus}{(theta) (document)}{M};%

\node[latent, right=4 of b,scale=1.4](b1){$\tilde{\beta}^{t}$};%
\node[latent, left=1 of b1,scale=1.4](p1){$\beta^{t}$};%
\node[latent, above=0.8 of p1,scale=1.4](wg1){$\tilde{W}^{t}$};%
\node[latent, above=0.8 of b1,scale=1.4](g1){$h^{t}$};%

\node[obs, below=1.25 of b1,scale=1.4](w1){$w$};%
\node[latent, left=0.5 of w1,scale=1.4](z1){$z$};%
\node[latent, below=0.8 of z1,scale=1.4](theta1){$\theta$};%
\node[obs, left=0.7 of theta1](a1){$\alpha$};%
\node[obs, above = 7 of a1,scale=1.4](eta) {$G,X$};%
\plate[inner sep=0.1cm,yshift=0cm]{document1}{(z1) (w1)}{N};%
\plate[inner sep = 0.25cm,yshift=0.1cm]{corpus1}{(theta1) (document1)}{M};%

\edge{a} {theta};%
\edge{theta} {z};%
\edge{z}{w};%
\draw[double, ->](p)--(b);%
\edge{b}{w};%
\draw[double,->](eta)--(g);%

\edge{a1} { theta1};%
\edge{theta1} {z1};%
\edge{z1}{w1};%
\draw[double,->](p1)--(b1);%
\draw[double,->](wg1)--(g1);%
\draw[double,->](wg)--(g);%
\draw[double,->](g)--(b);%
\draw[double,->](g1)--(b1);%
\edge{b1}{w1};%
\draw[double,->](eta)--(g1);%
\draw[->] (wg) to [out =30, in =150](wg1);%
\draw[->] (p) to [out =30, in =150](p1);%
\end{tikzpicture}
}
\caption{The graphical representation of GCTM. Single lines demonstrate stochastic processes while double lines show deterministic processes}
\label{fig:glda}
\end{figure}

\subsection{Proposed Model}


In this subsection, we describe how to integrate GCN \cite{kipf2017semi} into LDA to exploit a knowledge graph. Given prior knowledge of graph form $G = (V,E)$ where $V$ is a set of nodes which are words in the vocabulary and $E$ is a set of edges which encode particular relationships between words, we use graph convolutional networks with $L$ layers to learn the representation of nodes (words) in the graph. In detail, let $A$ ($A \in \mathbb{R}^{V \times V}$) be the adjacency matrix of $G$ and $X$ ($X \in \mathbb{R}^{V \times M}$) be a feature matrix in which each row $X_i$ ($i \in \{1,...,V\}$) is an $M$-dimensional feature vector of each word $i$. In GCN, each layer can encode neighbourhood relationship to learn a representation for all nodes in the graph. The representation $h_l$ of the nodes in layer $l$ is computed as follows:
\begin{align*}
h_l = f\left(\tilde{D}^{-\frac{1}{2}}\tilde{A}\tilde{D}^{-\frac{1}{2}} (h_{l-1}W_l+b_l) \right)
\end{align*}      
where $ \tilde{A} = A + I_V $ ($I_V$ is the identity matrix), $\tilde{D}_{ii} = \sum_j \tilde{A}_{ij}$, $\tilde{W}_{l} =\{W_l,b_l\}$ is the weight matrix of parameters.  $h_0$ is the feature matrix $X$ and the activation function $f$ is usually ReLU function. In the output layer, the dimension of word representation is set by $K$ in order to fit the number of topics $K$ in LDA ($h_L$ is a $V \times K$ matrix and each $K$-dimensional vector $h_{Lv}$ is the representation of word $v$). Then, we use a transpose operator on $h_L$ to be able to integrate with topic matrix $\beta$ of size $K \times V$.  This deterministic process is summarized concisely as: $h = GCN(h_0,G;\tilde{W})$ where $h_0$ is an input, $\tilde{W}$ is a weight matrix of GCN, and $h$ is an output ($h$ is a transpose matrix of $h_L$).  

Moreover, we need a mechanism to connect $\beta$ and $h$. In general, this mechanism can be represented by a function $F(\beta,h;\rho)$ where $\beta$ and $h$ are inputs, and $\rho$ is  parameter. For simplicity, we use a linear function to combine $\beta$ and $h$ on each topic $k$. Then, topic distribution $\tilde{\beta}_k$ is generated by using the softmax function. In detail, for each topic $k$ ($k \in \{1,...,K\}$), 
\begin{align*}
\tilde{\beta}_k = softmax(\rho_k\beta_k + (1 - \rho_k)h_k)
\end{align*}
where $\rho_k$ is a scalar to balance $\beta_k$ and $h_k$. In training, we must learn $\beta$, $\tilde{W}$, and $\rho$.  

For data streams, we base on the recursive Bayesian approach to keep the impact of learnt model from the previous minibatch to the current one. We assume that the models at two consecutive minibatches are connected by the following transition:
\begin{align*}
p(\beta^t|\beta^{t-1},\sigma_{\beta}) &= \mathcal{N}(\beta^t; \beta^{t-1},\sigma_{\beta}^2I)\\
p(\tilde{W}^t|\tilde{W}^{t-1},\sigma_w) &= \mathcal{N}(\tilde{W}^t; \tilde{W}^{t-1},\sigma_w^2I)
\end{align*}
where $\sigma_{\beta}$ and $\sigma_{w}$ are parameters that relate to the change of $\beta^t$ and $\tilde{W}^t$ around $\beta^{t-1}$ and $\tilde{W}^{t-1}$ respectively.

The generative process (Figure\ref{fig:glda}) of documents in a minibatch $t$ is described explicitly as below:
\begin{enumerate}
\item Draw $\tilde{W}^{t} \sim \mathcal{N}(\tilde{W}^t; \tilde{W}^{t-1},\sigma_w^2I)$
\item Calculate $h^t = GCN(h_0,G;\tilde{W}^t)$
\item Draw $\beta^t \sim \mathcal{N}(\beta^t; \beta^{t-1},\sigma_{\beta}^2I)$
\item Calculate topic distribution:
\begin{align}
\tilde{\beta}^t = softmax(\rho^{t} \beta^{t} + (1 - \rho^t) h^t))
\label{eq:beta}
\end{align}
\item For each document $d$:
\begin{enumerate}
\item Draw topic mixture: $\theta_d  \sim Dirichlet( \alpha )$
\item For the $ n^{th} $ word of $d$:
\begin{enumerate}
\item Draw topic index: $ z_n  \sim Multinomial(\theta_d) $
\item Draw word: $w_n  \sim Multinomial(\tilde{\beta}^t_{z_n}) $
\end{enumerate}
\end{enumerate} 
\end{enumerate}

\subsection{Learning GCTM}

At a minibatch $t$, new documents arrive and are collected in a set of $D$ documents. 
 The posterior 
\begin{align*}
    p(\beta^t, \tilde{W}^t| D^t, \beta^{t-1}, \tilde{W}^{t-1}, G, X, \rho^t, \alpha, \sigma_{\beta}, \sigma_{w})
\end{align*}
is expressed as follows:


\begin{align}
	& \log p(\beta^t, \tilde{W}^t| D^t, \beta^{t-1}, \tilde{W}^{t-1}, G, X, \rho^t, \alpha,\sigma_{\beta}, \sigma_{w}) \nonumber \\ 
	&\propto \log p(\beta^t, \tilde{W}^t, D^t|\beta^{t-1}, \tilde{W}^{t-1}, G, X, \rho^t, \alpha, \sigma_{\beta}, \sigma_{w}) \nonumber \\ 
	&\propto \log p(\tilde{W}^t|\tilde{W}^{t-1}, \sigma_{w}) + \log p(\beta^t|\beta^{t-1}, \sigma_{\beta}) \nonumber \\
	&+ \log p(D^t| \beta^t,\tilde{W}^t, G, X, \rho^t, \alpha) = L \label{Eq_Pos}
\end{align}

We learn GCTM based on maximizing $L$ (Eq \ref{Eq_Pos}). We apply $\tilde{\beta^{t}} = softmax(\rho^{t}\beta^{t-1} + (1 - \rho^t)GCN(X,G;\tilde{W}^t))$ into Eq \ref{Eq_Pos}: 

\begin{align*}
L &= \log p(\tilde{W}^t|\tilde{W}^{t-1}, \sigma_{w}) + \log p(\beta^t|\beta^{t-1}, \sigma_{\beta}) \\
&+ \log p(D^t| \tilde{\beta}^{t}, \alpha)\\ 
& =  -\frac{1}{2\sigma_{\beta}^2} ||\beta^t - \beta^{t-1}||_{F}^2 - \frac{1}{2\sigma_{w}^2} || \tilde{W}^t - \tilde{W}^{t-1} || _F^2 \\
&+  \log p(D^t| \tilde{\beta}^{t}, \alpha)
\end{align*}

Because $p(D^t| \tilde{\beta}^{t}, \alpha)$ is intractable to compute, we use variational inference as in  \cite{blei2003latent} to do inference local variables $z$ and $\theta$. After applying Jensen inequality, we get evidence lower bound (ELBO):


\begin{align*}
L &= -\frac{1}{2\sigma_{\beta}^2} ||\beta^t - \beta^{t-1}||_{F}^2 - \frac{1}{2\sigma_{w}^2} || \tilde{W}^t - \tilde{W}^{t-1} || _F^2 \nonumber\\
&+  \log \int \sum_{z} \frac{p(D^t, \theta, z| \tilde{\beta}^{t}, \alpha)}{q(\theta,z)} q(\theta,z) d \theta \nonumber \\ 
& \geq -\frac{1}{2\sigma_{\beta}^2} ||\beta^t - \beta^{t-1}||_{F}^2 - \frac{1}{2\sigma_{w}^2} || \tilde{W}^t - \tilde{W}^{t-1} || _F^2 \nonumber \\
&+  \int \sum_{z} q(\theta,z) \log  \frac{p(D^t, \theta, z| \tilde{\beta}^{t}, \alpha)}{q(\theta,z)}  d \theta \nonumber \\ 
& \geq -\frac{1}{2\sigma_{\beta}^2} ||\beta^t - \beta^{t-1}||_{F}^2 - \frac{1}{2\sigma_{w}^2} || \tilde{W}^t - \tilde{W}^{t-1} || _F^2 \\
&+ E_{q(\theta,z)}[\log p(D^t, \theta, z| \tilde{\beta}^{t}, \alpha)] - E_{q(\theta,z)}[\log q(\theta, z)] \\
&= ELBO	
\end{align*}
where $q(\theta, z)$ is a factorized variational distribution:
\begin{align}
q(\theta, z) = \prod_{d=1}^{D} \left( Dirichlet(\theta_d|\gamma_d) \prod_{n=1}^{N_d} Multinomial (z_{dn}|\phi_{dn}) \right)
\end{align} 
$\gamma$ and $\phi$ are variational parameters. When $\alpha$, $\sigma_{\beta}$ and $\sigma_{w}$ are fixed, we maximize ELBO with respect to local parameters ($\gamma$ and $\phi$) and global parameters ($\rho^t$, $\beta^t$, and $\tilde{W}^t$). According to \cite{blei2003latent}, the update equations of local parameters are:

\begin{align}
\gamma_{dk} &\leftarrow \alpha_k + \sum_{n=1}^{N_d}\phi_{dnk}\qquad \text{for } k=1,...,K \label{eq:gamma}  \\
\phi_{dnk} &\propto \exp (E_q[\log \theta_{dk}] + \sum_{v=1}^V I[w_{dn}=v] \log \tilde{\beta}_{kv}) 
\label{eq:phi}
\end{align}
where $I[\cdot]$ is an indicator function and $E_q[\log \theta_{dk}] = \psi(\gamma_{dk})-\psi (\sum_{k=1}^K (\gamma_{dk}))$ ($\psi$ is a digamma function). 

Regarding global parameters, we extract the part of ELBO w.r.t $\rho^t$, $\beta^t$, and $\tilde{W}^t$:
\begin{align}
&ELBO(\rho^t,\beta^t,\tilde{W}^t) \nonumber \\
& =-\frac{1}{2\sigma_{\beta}^2} ||\beta^t - \beta^{t-1}||_{F}^2 - \frac{1}{2\sigma_{w}^2} || \tilde{W}^t - \tilde{W}^{t-1} || _F^2 \nonumber \\
&+\sum_{d=1}^M \sum_{n=1}^{N_d} \sum_{v=1}^V \sum_{k=1}^K I(w_{dn}=v) \phi_{dnk} \log (\tilde{\beta}^t_{kv})
\label{eq:elbo}
\end{align}

where $\tilde{\beta}^t = softmax(\rho^{t}\beta^{t-1} + (1 - \rho^t)GCN(X,G;\tilde{W}^t))$. We use Adam \cite{kingma2014adam} to maximize $ELBO(\rho^t,\beta^t,\tilde{W}^t)$.

The whole learning process of GCTM is presented in Algorithm \ref{algo:GCNLDA}.

\begin{algorithm}[tp]
    \caption{Learning GCTM}
	\label{algo:GCNLDA}
\begin{algorithmic}
\REQUIRE{ Graph G, hyper-parameter $\alpha$, data sequence $\{D^1,D^2,...\}$}
\ENSURE{$\tilde{W}, \beta, \rho$ \\}
Initialize $\tilde{W}^0, \beta^0$ randomly\\
\FOR {minibatch $t$ with data $D^t$} 
\STATE {Compute $\tilde{\beta}$ by Equation (\ref{eq:beta})}
\FOR{each document $d$ in $D^t$}
\STATE {Infer $\gamma_d$ and $\phi_d$ by iteratively updating (\ref{eq:gamma}) and (\ref{eq:phi}) until convergence}
\ENDFOR\\
\STATE{Update $\tilde{W}^t, \beta^t, \rho^t$ by using Adam \cite{kingma2014adam} to maximize (\ref{eq:elbo})}
\ENDFOR
\end{algorithmic}
\end{algorithm}

\subsection{Discussion}
In this subsection, we discuss the advantages of GCTM and compare it with other methods. GCTM can well exploit an external knowledge graph for data streams. Therefore, we discuss some aspects of this topic. 

First, GCN, which is an effective model to encode relationships between edges in a graph, can learn graph embedding to fit the form of topic matrix in LDA. Therefore, our method can utilize the graph embedding to enrich information for learning topics better. To the best of our knowledge, this is the first work which can exploit a prior knowledge graph for LDA in the streaming environment. Meanwhile, almost existing streaming methods ignore prior knowledge; and KPS aims to use but is limited to prior knowledge of vector form.   
  
Second, in each minibatch, our method provides a mechanism to automatically balance old knowledge (that is obtained from the previous minibatch) and a prior knowledge graph. Meanwhile, KPS \cite{kpspakdd} must manually control the impact of prior knowledge in each minibatch, and it is difficult to tune this impact in streaming environments. 

Third, our method can deal with concept drift well when data arrives continually. Using external knowledge that covers or relates to new concepts is an effective solution to handle concept drift. However, it is difficult to guarantee that prior knowledge contains information about new concepts. Fortunately, this is possible with a knowledge graph such as Wordnet or a graph trained on a big dataset. Especially, when new topics occur, a set of new words can be used to describe them. However, the words and their relations are also included in the knowledge graph. As a result, exploiting the graph helps our method to learn new topics in new arriving documents. On the other hand, many streaming methods suffer from concept drift because they only use old knowledge learnt from the previous minibatch as prior in the current minibatch. It means that emphasizing the old knowledge prevents the model from adapting to new data. HPP \cite{masegosa17power} also has a mechanism to combine old knowledge and initial prior. It deals well with concept drift in cases that the prior is good enough and the mechanism helps to forget the old knowledge. In our work, we also use a similar mechanism, but exploit better external knowledge. 

Finally, our method learns both GCN and LDA simultaneously in the streaming environment. More generally, it can be extended to train a hybrid model of a neural network and a probabilistic model for data streams.

\section{Evaluation}
In this section, we conduct intensive experiments to evaluate the performance of our method in terms of log predictive probability and topic coherence on several datasets (both short and regular text datasets) in the streaming environment. We also examine how our method deals with concept drift. Finally, we investigate the sensitivity of our method w.r.t hyperparameters. 

\subsection{Datasets and Baselines} 


\begin{table}
\centering
\caption{Some statistics about the datasets.}
\begin{tabular}{lllll}
Dataset & Vocab & Training & Evaluation & words/doc \\ \hline
Agnews   & 32,483 & 110,000 & 10,000 & 24.9 \\ 
TMN & 11,599 & 31,604 & 1,000 & 24.3 \\ 
NYT-title & 46,854 & 1,664,127 & 10,000 & 5.0 \\ 
Yahoo-title & 21,439 & 517,770 & 10,000 & 4.6 \\ 
Agnews-title   & 15,936 & 108,400 & 10,000 & 4.9 \\ 
TMN-title & 2,823 & 26,251 & 1,000 & 4.6 \\ 
Irishtimes & 28,816 & 1,364,669 & 10,000 & 5.0 \\ 
Twitter & 35072 & 1247321 & 10000 & 6.2 \\ 
\end{tabular}
\label{data}
\end{table}

We conduct experiments on $6$ short text datasets (NYT-title \footnote{http://archive.ics.uci.edu/ml/datasets/Bag+of+Words}, Yahoo-title\footnote{https://answers.yahoo.com/}, TagMyNews-title (TMN-title), Irishtimes\footnote{https://www.kaggle.com/therohk/ireland-historical-news/}), Agnews-title, Twitter\footnote{http://twitter.
com/} and $2$ regular text datasets (Agnews\footnote{https://course.fast.ai/datasets}, TagMyNews (TMN)\footnote{http://acube.di.unipi.it/tmn-dataset/}). The Yahoo-title and Twitter datasets \cite{mai2016enabling,tuan2020bag} are crawled from a forum and a social network respectively, therefore they often contain noisy texts. The datasets are preprocessed with some steps such as: tokenizing, removing stopwords and low-frequency words (which appear in less than 3 documents) to build the corresponding vocabularies, and removing extremely short documents (less than 3 words). The statistics of these datasets are described in Table \ref{data}. Experimenting on the short text corpora, in which each document contains about $5$ words, helps us to examine the role of a knowledge graph in case of short and sparse data. 

\paragraph{Knowledge graphs:}

In these experiments, we exploit external knowledge which is derived from both human knowledge (Wordnet\footnote{https://Wordnet.princeton.edu/}) and a pre-trained model (Word2vec\footnote{http://nlp.stanford.edu/projects/glove/}) on a big dataset. Wordnet and Word2vec are used to create $2$ knowledge graphs respectively. In terms of building the Wordnet graph, for each word in the vocabulary of each dataset, we get all words that have either synonym or antonym relationships with it from Wordnet to create a set of its word neighbors. However, in order to avoid a big graph, we remove neighbors that are out of vocabulary. Then, an edge is built based on neighbor relation and the weight of each edge is the Wu-Palmer similarity of the corresponding pair of words.  We emphasize that we take neighbors with all different meanings for each word. Therefore, although concept drift happens or a word is used in a different meaning from previously appearing meanings, the Wordnet graph includes this meaning to enrich a topic model. For the other graph, we base on Word2vec to compute cosine similarity between a pair of words in the vocabulary. Then, for each word, we select the top $200$ words with highest similar score to build a graph. The $2$ graphs are used as prior knowledge for GCTM. 

First, we ignore node features to focus on evaluating the impact of a knowledge graph in streaming environments. It means that $X$ is set to be the identity matrix $I_V$. Then, we investigate the combination of both a knowledge graph from Wordnet and node features from Word2vec to enrich a topic model.  

\paragraph{Baselines:}
We use $3$ state-of-the-art baselines to learn LDA from data streams in comparison with our method. We briefly describe these methods as follows:
\begin{itemize}
\item Population variational Bayes (PVB) \cite{populationdis} uses stochastic natural Gradient ascent to maximize the expectation of the likelihood of data. 
\item Streaming variational Bayes (SVB) \cite{streamvb} bases on recursive Bayesian approach. SVB can only use external knowledge encoded in the prior at the first minibatch, then ignores it in the next minibatches.
\item Power prior (SVB-PP)\footnote{Due to requiring non-trivial efforts, SVB-HPP is not included in this paper. However, the original work \cite{masegosa17power} showed that if SVB-PP is tuned well, it is often comparable to SVB-HPP.}\cite{masegosa17power} is an extension of SVB. It can exploit the original prior distribution through all minibatches and provides a mechanism to control the impact of the prior in each minibatch. 
\item GCTM-WN: GCTM uses a knowledge graph from Wordnet.
\item GCTM-W2V: GCTM exploits a knowledge graph from Word2vec.
\end{itemize} 

The same hyperparameters in all methods are set the same. In detail, we set the hyperparameter of Dirichlet distribution $\alpha=0.01$ for topic proportion of each document, the number of topics $K=50$ for {Agnews, Agnews-title, TMN, TMN-title} and $K=100$ for {Yahoo-title, NYT-title, Irishtimes}. We note that the baselines cannot exploit a prior knowledge graph, they only use a Dirichlet prior with a hyperparameter $\eta=0.01$ for each topic as in the original papers. For other hyperparameters, we use grid search to determine the best hyperparameter for each method on each dataset. In detail, the range of each hyperparameter is set as follows: the multiple power prior $\rho \in \{0.6, 0.7, 0.8, 0.9, 0.99\}$ for SVB-PP,  the population size $S$ in $\{10^3, 10^4, 10^5, 10^6\}$ and the forgetting factor $\kappa$ in $\{0.7, 0.8, 0.9,0.99\}$ for PVB, and variance $\sigma_{\beta}=\sigma_w =\sigma \in \{0.1, 1, 10\}$, the number of GCN layers $L=2$ for GCTM.  We list the best hyperparameters of the methods from grid search in appendix C.

\paragraph{Performance measure:}
We use $2$ measures to evaluate the methods: Log predictive probability (LPP) \cite{lpp} which considers the generalization of a model and Normalized  pointwise  mutual  information  (NPMI) \cite{lau2014machine} which exams the coherence of topics. We measure the LPPs of the methods after every minibatch. However, due to computing on all documents of each dataset, NPMI is only measured after finishing the whole training process. We describe these measures in appendices A and B.

\subsection{Experiments on datasets with fixed batchsize}
\begin{figure*} 
\begin{center}
\includegraphics[width=1\textwidth]{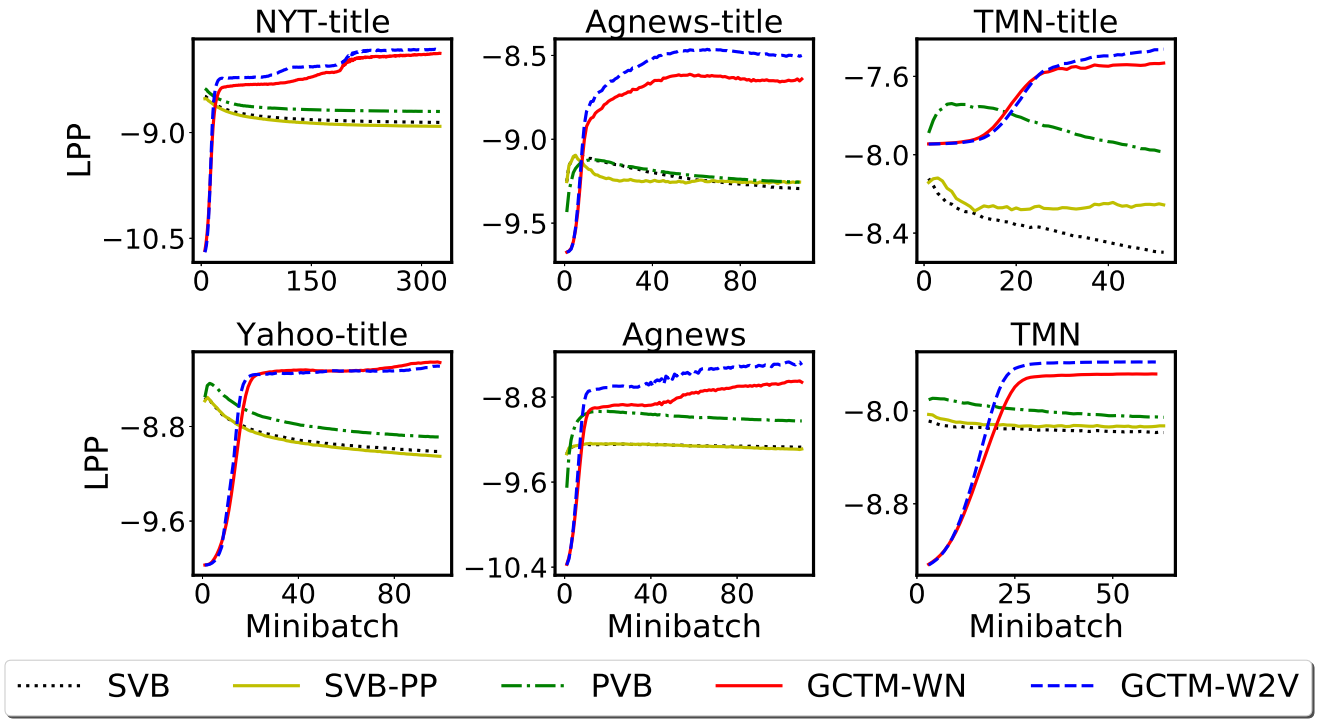} 
\end{center}
\caption{Performance of the methods in terms of generalization as learning from more data. Higher is better.}
\label{lpp}
\end{figure*}

%

\begin{table*}[width=0.9\textwidth,cols=6]
  \caption{Performance of the methods in terms of topic coherence. Higher is better.}
  \begin{tabular*}{\tblwidth}{@{} LLLLLL@{} }
   \toprule
    Dataset & GCTM-WN & GCTM-W2V & SVB & SVB-PP & PVB \\ 
   \midrule
    Agnews   & 0.287 & 0.263 & 0.005 & 0.005 & 0.018\\
    Agnews-title & -0.026 & -0.067 & -0.114 & -0.111 & -0.107\\ 
    TMN  & 0.0791 & 0.073 & -0.015 & -0.054 & -0.019 \\ 
    TMN-title & 0.032 & 0.021 & -0.103 & -0.105  & -0.090 \\ 
    NYT-title & 0.266 & 0.246 & -0.069 & -0.065 & -0.069 \\ 
    Yahoo-title & 0.171 & 0.173 & -0.087 & -0.088 & -0.076 \\ 
   \bottomrule
  \end{tabular*}
  \label{npmi}
\end{table*}


Due to the lack of time information on almost datasets (except Irishtimes dataset), we simulate the streaming environment by following experimental designs in \cite{streamvb,populationdis}. We conduct experiments with the scenarios on $6$ datasets (TMN, TMN-title, Agnews, Agnews-title, Yahoo-title, and NYT-title). In each dataset, we randomly select a holdout test set (which contains documents with more than or equal to $5$ words) and then shuffle the remaining documents and divide them into minibatches with fixed batchsize for training. Based on the size of each dataset, we set batchsize to $500$ for {TMN, TMN-title}, $1000$ for {Agnews, Agnews-title}, and $5000$ for {Yahoo-title, NYT-title}. The information of training and test sets is described in Table \ref{data}. 

In terms of LPP, Figure \ref{lpp} shows the experimental results. We have some noticeable observations from these results. First, both GCTM-WN and GCTM-W2V significantly outperform the baselines. Providing an external knowledge graph from Wordnet or Word2vec is the main reason why the GCTM-based methods achieve better performances than the baselines which do not exploit prior knowledge. Second, both GCTM-WN and GCTM-W2V are inferior to the baselines in a few beginning minibatches on NYT-title and Yahoo-title datasets, while they need more minibatches to catch up with the baselines on the remaining datasets. Due to having to learn a lot of parameters in graph convolutional networks, the GCTM-based methods need more data to learn the model. Moreover, the differences of batchsize among datasets lead GCTM-WN and GCTM-W2V to require the different numbers of minibatches to overcome the baselines. Third, the performances of the baselines only increase in a few beginning minibatches, then gradually decrease on short text datasets. It means that the baselines deal badly with short texts even though the data is big. In contrast, the GCTM-based methods with external knowledge can work well on short texts. Finally, in comparison with the baselines, the improvements of the GCTM-based methods on the short text datasets (Agnews-title and TMN-title) are more remarkable than those on the regular text datasets (Agnews and TMN respectively). This provides convincing evidence of exploiting external knowledge for data streams.

Regarding NPMI, Table \ref{npmi} shows the experimental results. Both GCTM-WN and GCTM-W2V also outperform the baselines by noticeable margins. Because Wordnet and Word2vec, which encode the information of word semantic and local contexts, help LDA to learn coherent topics. The regular text datasets (Agnews and TMN) contain more information of word co-occurrence than the short ones, therefore, the methods work better on the regular datasets. Moreover, the GCTM-based methods also perform more significantly on the short text datasets.   

The different graphs from Wordnet and Word2vec have different impacts in terms of LPP and NPMI. It seems that the word-embeddings-based graph improves LDA slightly better than the Wordnet-based graph in terms of LPP on all the datasets (Figure \ref{lpp}). However, GCTM-W2V performs worse than GCTM-WN in terms of topic coherence (Table \ref{npmi}).

\subsection{Experiments on dataset with timestamp}
\begin{figure*} 
\begin{center}
\includegraphics[width=1\textwidth]{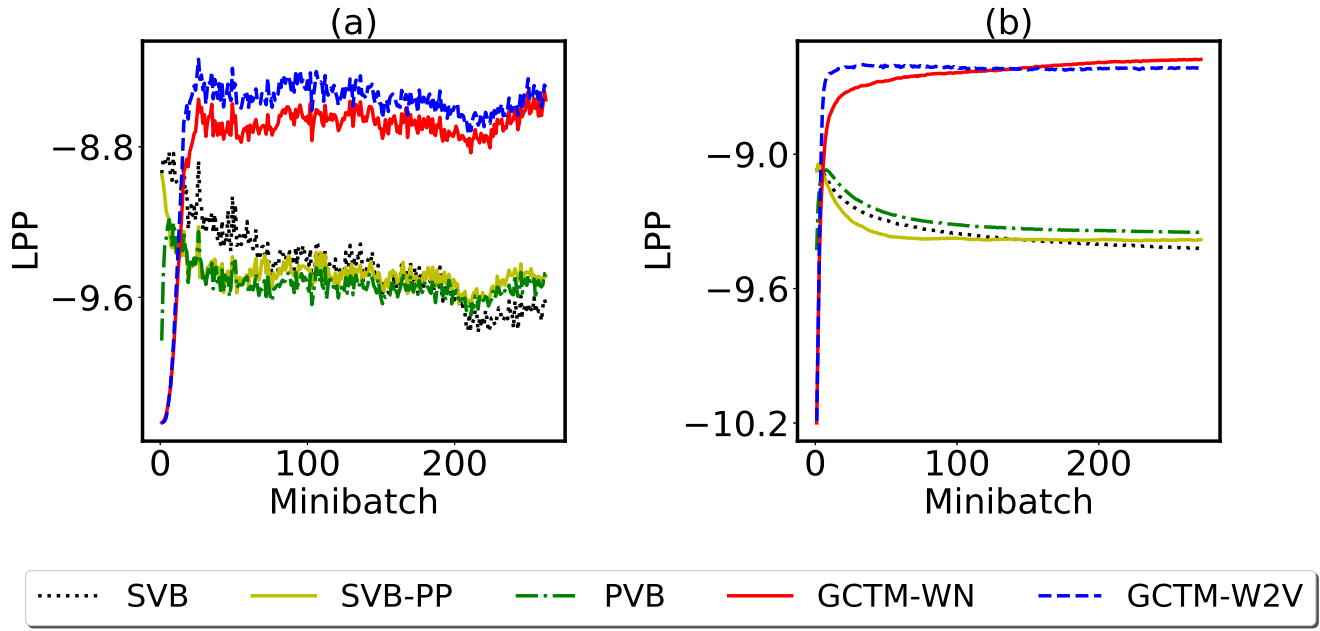} 
\end{center}
\caption{Performance of the methods on the Irishtimes dataset. While Figure (a) shows the results on the dataset with timestamp, Figure (b) reports the results on the dataset with fixed batchsize.}
\label{irishtimes}
\end{figure*}

%

\begin{table*}[width=.9\textwidth,cols=6]
  \caption{Performance of the methods in terms of topic coherence on the Irishtimes with both time stamp and fixed batchsize.}
  \begin{tabular*}{\tblwidth}{@{} LLLLLL@{} }
   \toprule
    Dataset & GCTM-WN & GCTM-W2V & SVB & SVB-PP & PVB \\
   \midrule
    Timestamp  &  0.127 & 0.124 & -0.068 & -0.083 & -0.082 \\ 
    Fixed batchsize & 0.002 & 0.002 & -0.068 & -0.072 & -0.065\\ 
   \bottomrule
  \end{tabular*}
  \label{npmi_irish}
\end{table*}

Since only the Irishtimes dataset has information about time, we only conduct experiments with timestamp on this dataset. We get the documents over period of each month to create a minibatch. GCTM is trained on a minibatch and the next minibatch is used to measure LPP. We use this scenario to evaluate the methods in a real streaming environment. We also conduct extra experiments with the previous scenario on this dataset to investigate the differences between the scenarios. For the extra experiments, we fix batchsize to $5000$ and the size of test set to $10000$.  In both scenarios, we evaluate NPMI on all documents in the dataset.

The LPP results are reported in Figure \ref{irishtimes}. While Figure \ref{irishtimes}(a) shows the results on the dataset with timestamp, Figure \ref{irishtimes}(b) illustrates the results on the dataset with fixed batchsize. It is obvious that the behaviours of lines in both scenarios are similar. In the timestamp scenario, the performances of the GCTM-based methods are significantly better than the baselines in terms of LPP.  However, the lines in Figure \ref{irishtimes}(a) are more curved than the ones in Figures \ref{irishtimes}(b). Since test set in each minibatch is the next one in the experiments with timestamp, the results are not as smooth as those in the other experiments with fixed holdout test set. Meanwhile, Table \ref{npmi_irish} shows that  the GCTM-based methods also achieve better NPMI results than the baselines in both the scenarios.

\subsection{Experiments on noisy data}

\begin{figure*} 
\begin{center}
\includegraphics[width=1\textwidth]{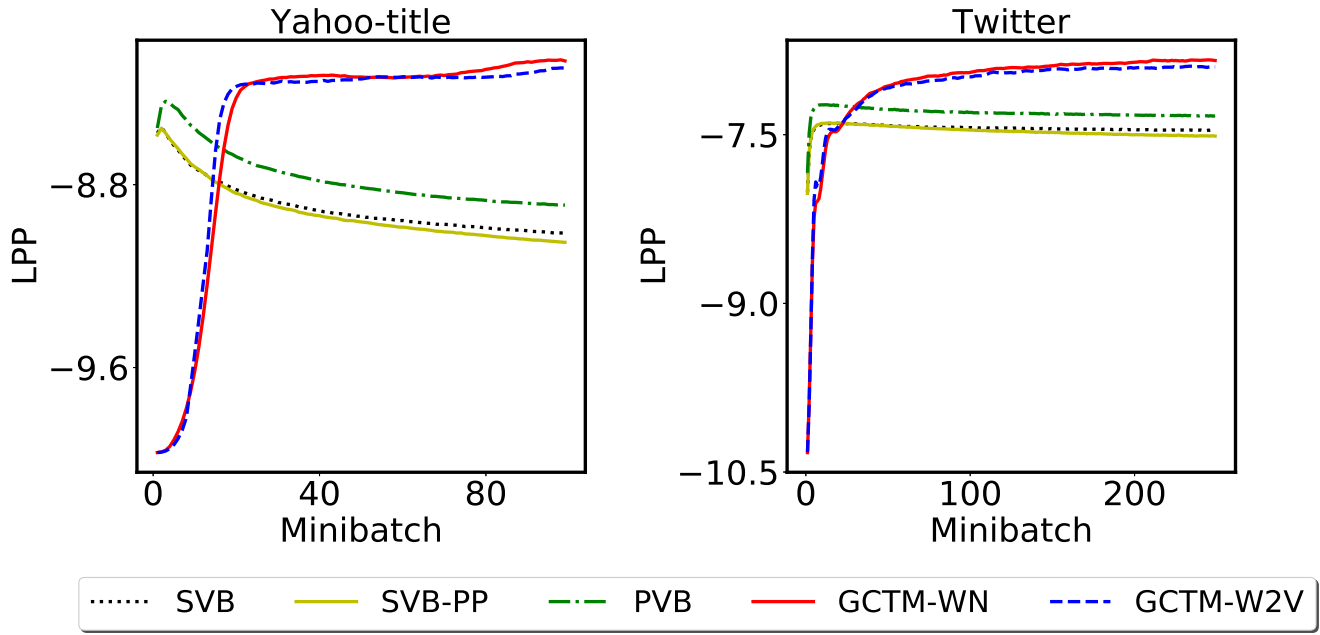}
\end{center}
\caption{Performance of the methods in terms of generalization when dealing with noisy texts. Higher is better.}
\label{fig:perplexities_noisydata}
\end{figure*}

%

\begin{table*}[width=.9\textwidth,cols=6]
 \caption{Performance of the methods in terms of topic coherence when dealing with noisy texts. Higher is better.}
  \begin{tabular*}{\tblwidth}{@{} LLLLLL@{} }
   \toprule
    Dataset & GCTM-WN & GCTM-W2V & SVB & SVB-PP & PVB \\
   \midrule
    Yahoo-title & 0.171 & 0.173 & -0.087 & -0.088 & -0.076 \\ \hline
    Twitter & -0.009 & -0.010 &-0.062 & -0.060 & -0.047 \\ \hline
   \bottomrule
  \end{tabular*}
  \label{npmi_irish}
\end{table*}
In this subsection, we consider how the methods deal with noisy texts. We conduct experiments on Yahoo-title and Twitter datasets. While the Twitter dataset is a collection of tweets from a social network\footnote{http://twitter.com/}, the Yahoo-title dataset is crawled from a question and answer forum\footnote{https://answers.yahoo.com/} where users freely post questions and others help to answer. Because texts from both the forum and social network are informal and contain noises, we can use them to evaluate performance of the methods when dealing with noisy data.

Figure \ref{fig:perplexities_noisydata} and Table \ref{npmi_noisydata} show the performances of the methods in terms of generalization and topic coherence respectively. It is straightforward to see that short and noisy texts not only rarely provide the baselines with enough word-occurrence information but also mislead them. As a result, the LPPs of the baselines decrease when more texts arrive after each minibatch. Moreover, NPMIs of the baselines do not obtain positive results. By using external knowledge graphs, both GCTM-WN and GCTN-W2V achieve better results than the baselines on both measures. These results provide experimental evidence why exploiting external knowledge in general and knowledge graph in particular is an effective solution to deal with noisy and short data.

\subsection{Experiments on dataset with concept drift and catastrophic forgetting}

\begin{figure*} 
\begin{center}
\includegraphics[width=1\textwidth]{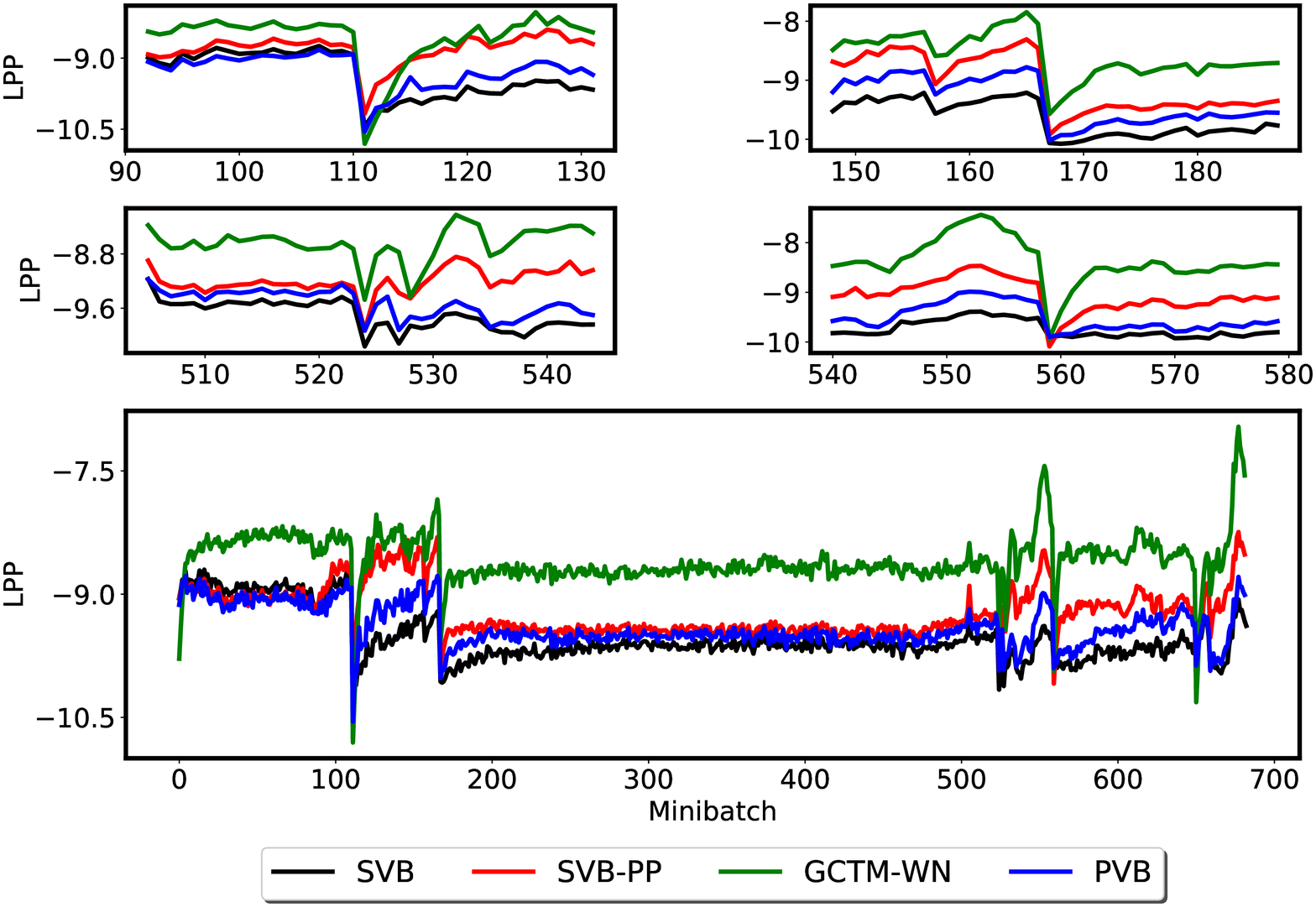} 
\end{center}
\caption{Performance of the methods when dealing with concept drift}
\label{concept_drift_kbl2}
\end{figure*}

\begin{figure*} 
\begin{center}
\includegraphics[width=0.95\textwidth]{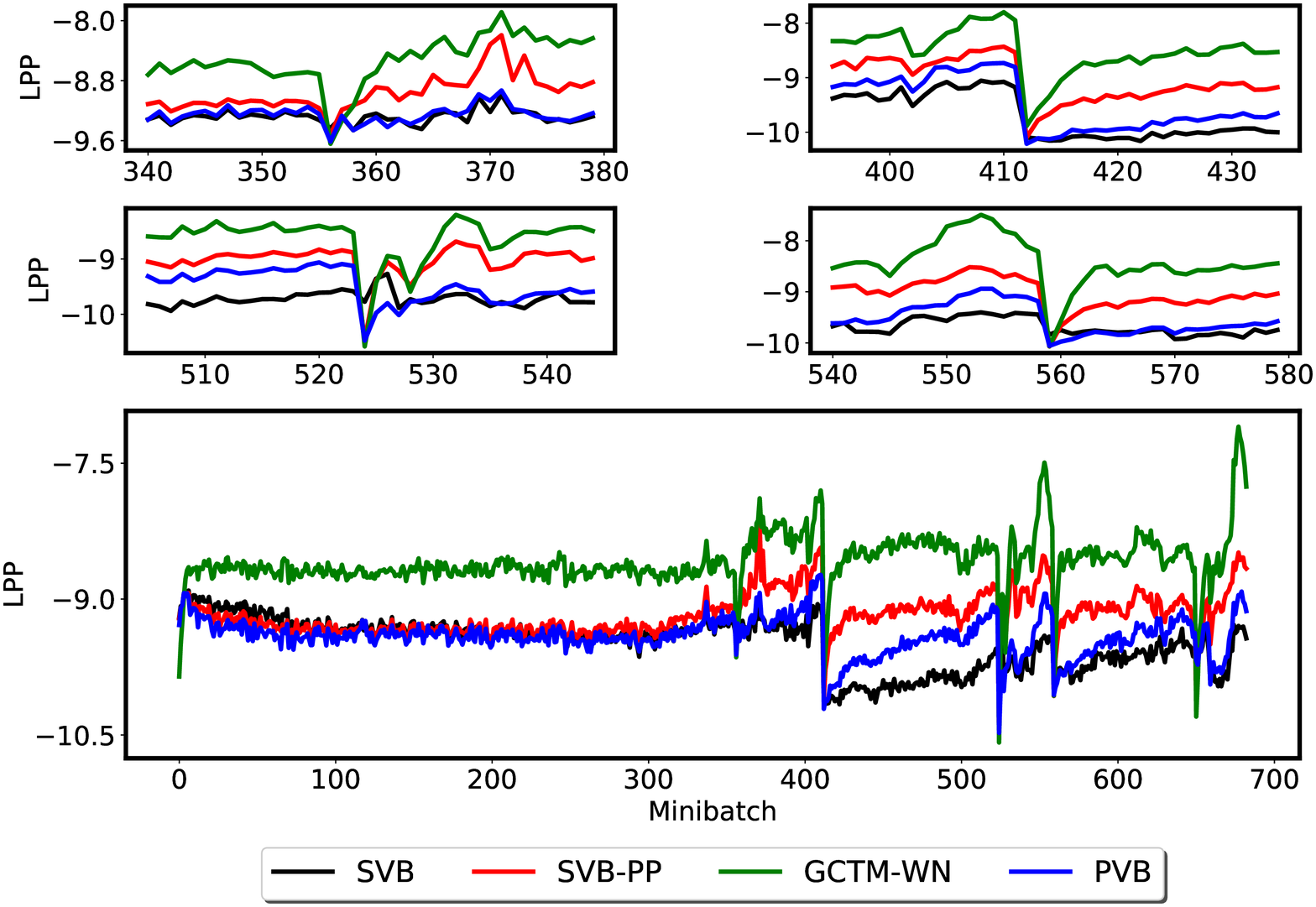} 
\end{center}
\caption{Performance of the methods when dealing with concept drift}
\label{concept_drift}
\end{figure*}

\begin{figure*} 
\begin{center}
\includegraphics[width=0.8\textwidth]{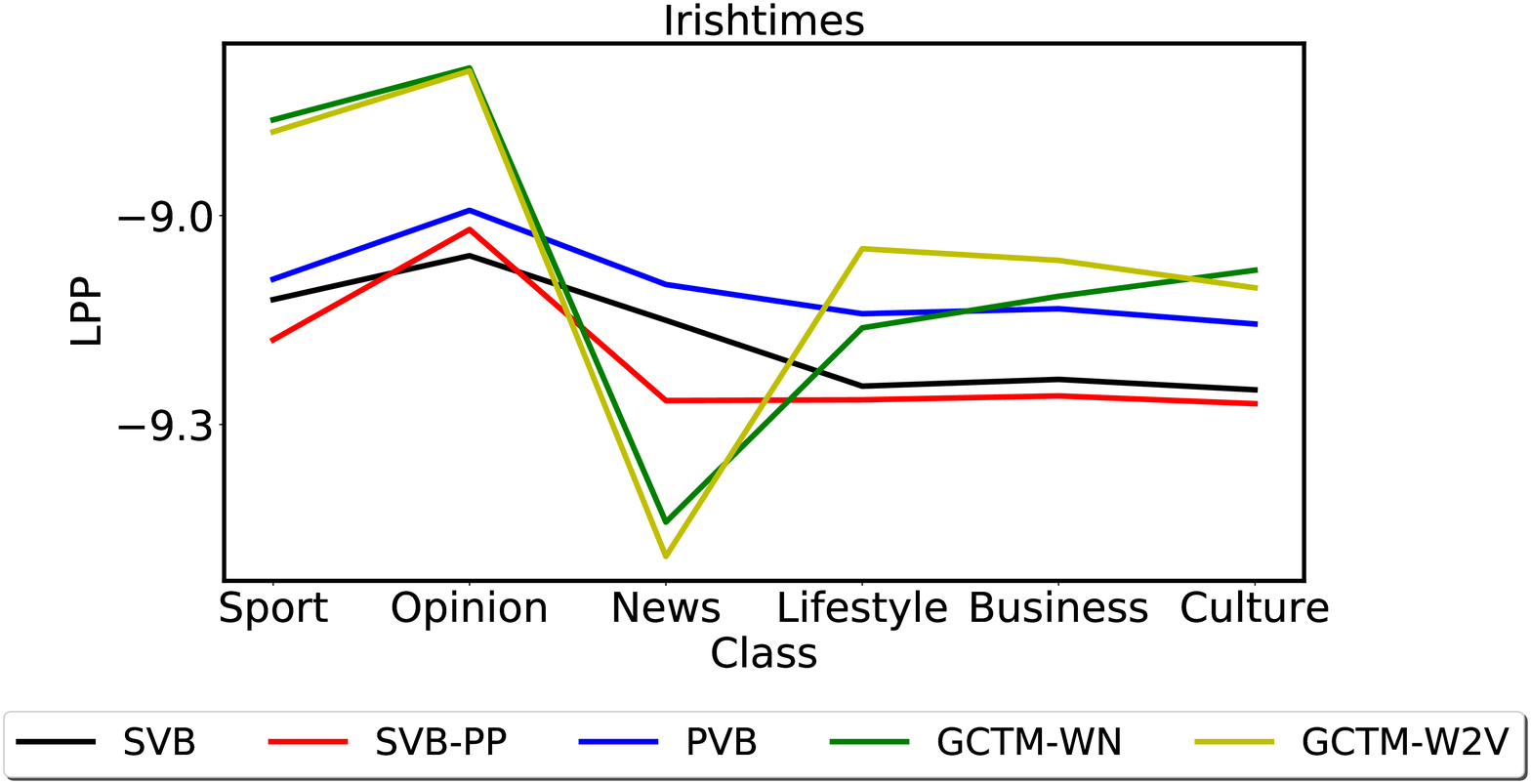} 
\end{center}
\caption{Catastrophic forgetting phenomenon after finishing training each class. LPP is averagely measured on hold-out test sets of the current and previous classes. Higher is better.}
\label{fig:forget1}
\end{figure*}
\begin{figure*} 
\begin{center}
\includegraphics[width=0.8\textwidth]{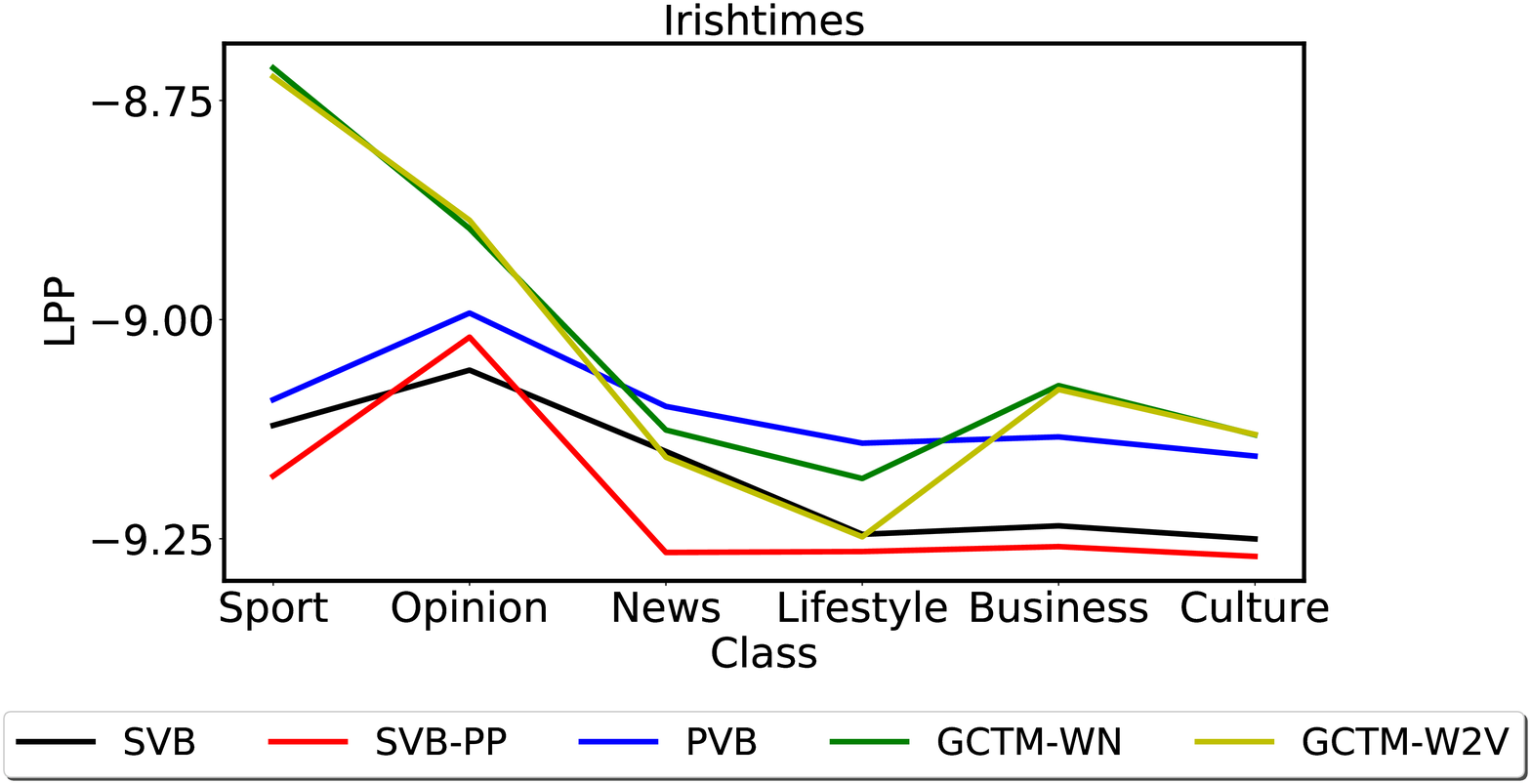} 
\end{center}
\caption{Catastrophic forgetting phenomenon after finishing training each class. LPP is averagely measured on hold-out test sets of the current and previous classes. Higher is better.}
\label{fig:forget3}
\end{figure*}

\textbf{Concept drift:} We design a scenario to evaluate the methods when dealing with concept drift. We simulate concept drift dataset on the Irishtimes dataset in which documents are categorized in $6$ classes whose labels are "News", "Opinion", "Sport", "Lifestyle", "Business", "Culture". We divide the dataset into minibatches with constraints as follows: Documents in the same minibatch have the same class label and the minibatches of the same class are used consecutively to train the model. Due to data imbalance in classes, batchsize is only set to $2000$. After training the model in a minibatch, we use the next one to measure LPP. In this scenario, concept drift arises when data changes from a particular class to a new one. It requires the model to adapt quickly to data of a new class. We conduct experiments with $2$ scenarios which are different in the order of labels. The number of texts in "News" is significantly bigger than those of other labels. We will change the order of this label. In detail, the first scenario uses the order of labels:  "News", "Opinion", "Sport", "Lifestyle", "Business", "Culture" while in the other scenario, labels are utilized sequentially in the following order: "Sport", "Opinion", "News", "Lifestyle", "Business", "Culture".

Figures \ref{concept_drift_kbl2} and \ref{concept_drift} illustrate the performances of the methods in the first and second scenarios respectively. Each figure includes $5$ subfigures: The main figure and $4$ small extra figures (which are extracted from the main figure to zoom in when concept drift happens). The main figures in both Figures \ref{concept_drift_kbl2} and \ref{concept_drift} show that GCTM-WN and SVB-PP achieve better results than PVB and SVB. Thanks to a balancing mechanism, both GCTM-WN and SVB-PP reduce the impact of old knowledge learnt from data of previous classes to work well on new data of the current class when concept drift happens. It is obvious that using a knowledge graph helps GCTM-WN outperform SVB-PP. Furthermore, the extra figures illustrate that the performances of the methods drop dramatically when concept drift arises. However, GCTM-WN increases significantly in a few minibatches, then remains stable. These results demonstrate that GCTM-WN can adapt quickly to concept drift.

\textbf{Catastrophic forgetting:} We examine the catastrophic forgetting phenomenon in which the methods forget the learnt knowledge when training on new data. We follow the measure of continual learning studies \cite{nguyen2018variational, kirkpatrick2017overcoming,ritter2018online} to consider the forgetting problem. In detail, we again use $2$ experimental scenarios in concept drift, however, we create a hold-out test set for each class. Each hold-out test set of each class consists of $2000$ texts. After finishing training all texts of a class, we calculate the average LPP on the hold-out test sets of the current and previous classes. The higher the average LPP of a method is, the better this method deals with the forgetting problem. 

Figure \ref{fig:forget1} and Figure \ref{fig:forget3} show the average LPPs of the methods after each class in both the scenarios. It is obvious that GCTM-WN and GCTM-W2V still achieve better results than the baselines at almost evaluation times. They are only inferior to the baselines a few times such as at the class "Sport" in Figure \ref{fig:forget1} and the classes "News" and "Lifestyle" in Figure \ref{fig:forget3}. Therefore, in both the scenarios, GCTM not only adapts more quickly to concept drift but also reduces more noticeably the catastrophic forgetting phenomenon in comparison with the baselines. It means that GCTM can deal better with the plasticity-stability dilemma than the baselines. However, it seems that GCTM deals with concept drift better than forgetting problem. The LPPs of GCTM in Figures \ref{concept_drift_kbl2} and \ref{concept_drift} are significantly higher than those in Figures \ref{fig:forget1} and \ref{fig:forget3}.

\subsection{Ablation studies}

In this subsection, we investigate the effectiveness of enriching Wordnet graph with node features from Word2vec as well as analyze the sensitivity of GCTM w.r.t hyperparameters. 

\subsubsection{Enriching Wordnet graph with node features from Word2vec in GCTM}

\begin{figure*} 
\begin{center}
\includegraphics[width=1\textwidth]{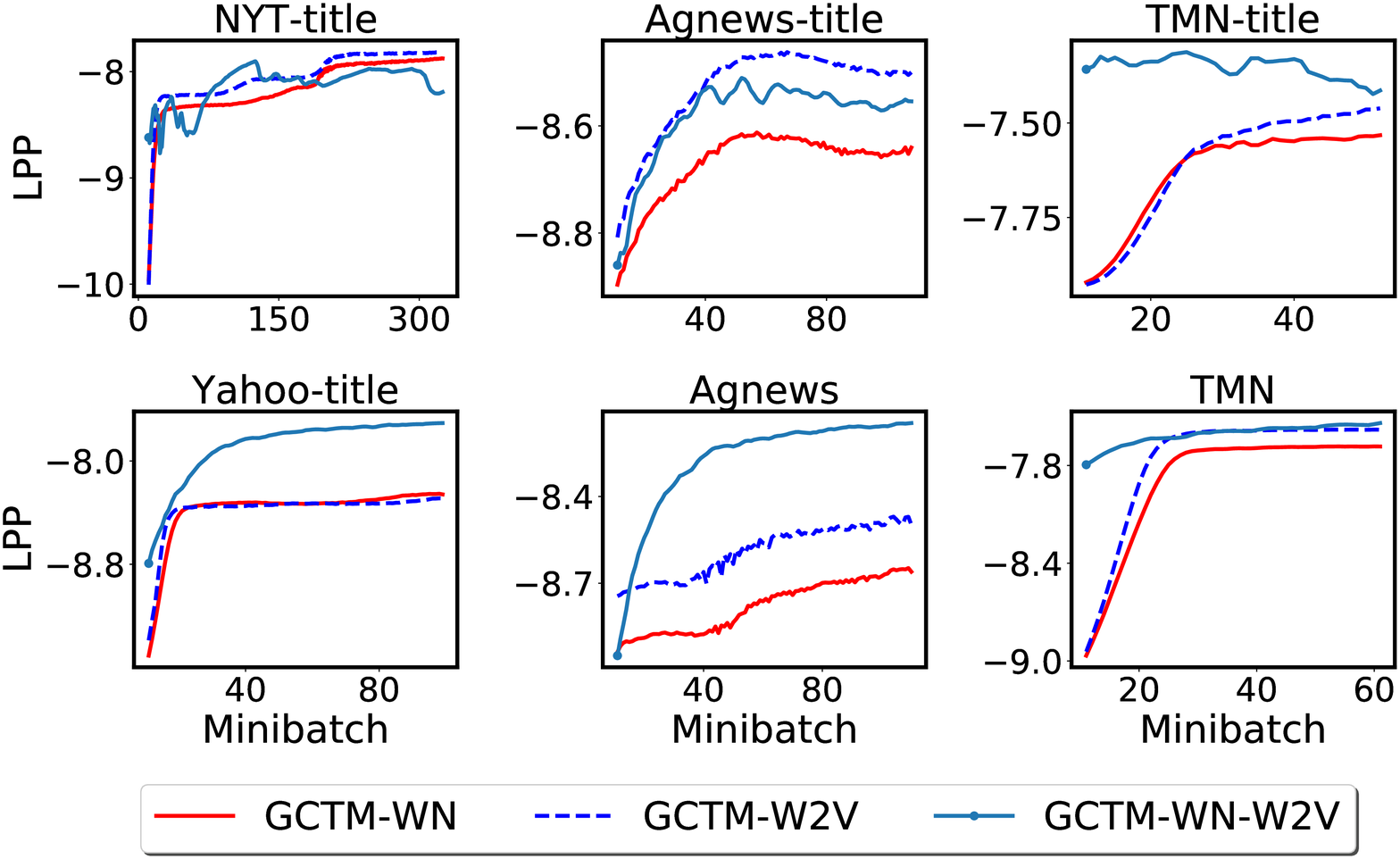} 
\end{center}
\caption{Performance of GCTM-WN-W2V with both the Wordnet graph and node features from Word2vec in comparison with GCTM-WN and GCTM-W2V. Higher is better.}
\label{fig:perplexities_w2v_wn}
\end{figure*}

We exploit both the Wordnet graph and the features of nodes from Word2vec to create GCTM-WN-W2V. We conduct experiments to compare this combination with GCTM-WN and GCTM-W2V which ignore node features. Figure \ref{fig:perplexities_w2v_wn} shows that GCTM-WN-W2V outperforms both GCTM-WN and GCTM-W2V with significant magnitudes in $3$ datasets: Yahoo-title, Agnews, and TMN-title. It achieves comparable results with the others in the TMN and NYT-title datasets. It is merely inferior to GCTM-W2V, but is superior to GCTM-WN on the Agnews-title. In particular, it is obvious that GCTM-WN-W2V is better than GCTM-WN. It means that exploiting good features of nodes in a knowledge graph can improve the effectiveness of GCTM.

\subsubsection{Sensitivity of GCTM w.r.t. hyperparameters}

\begin{figure*} 
\begin{center}
\includegraphics[width=0.98\textwidth]{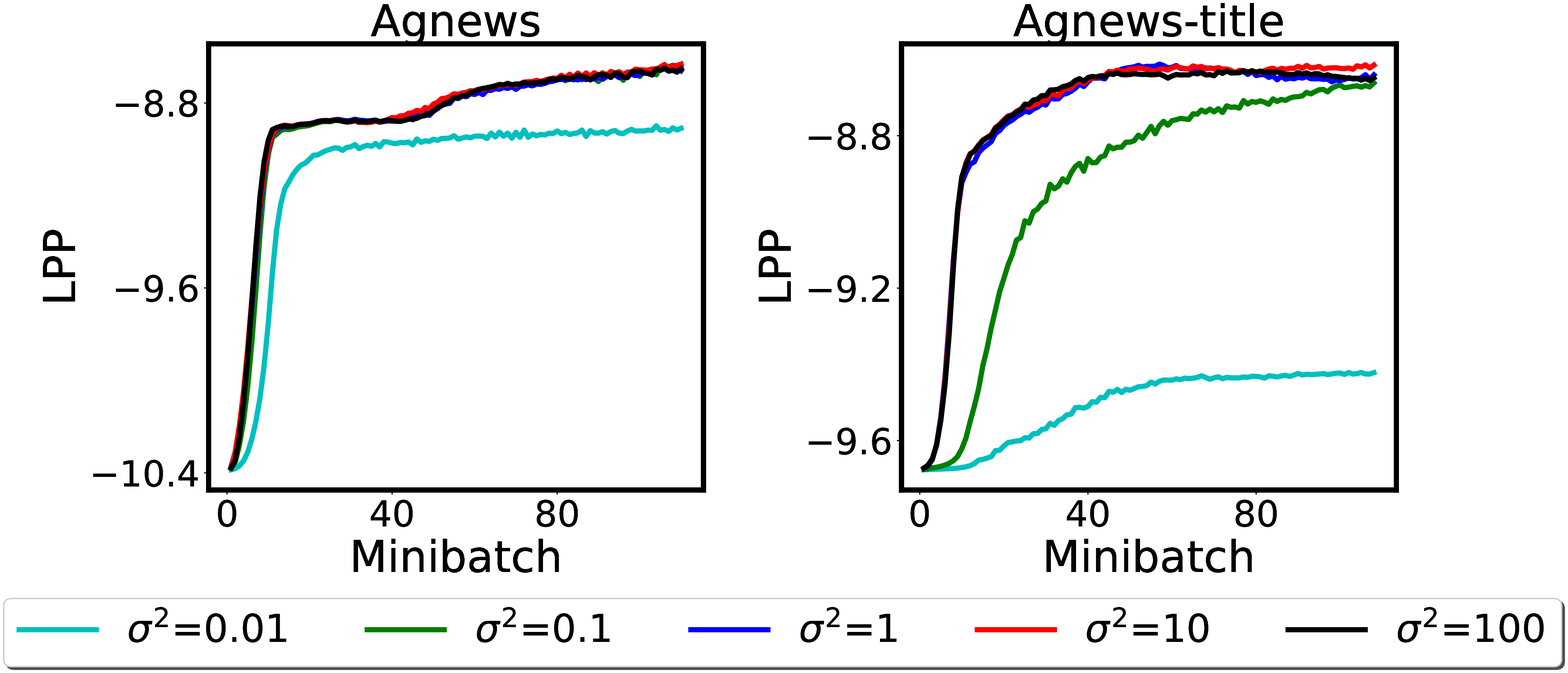} 
\end{center}
\caption{Sensitivity of GCTM-WN w.r.t $\sigma$}
\label{sens_sigma}
\end{figure*}

\begin{figure*} 
\begin{center}
\includegraphics[width=0.85\textwidth]{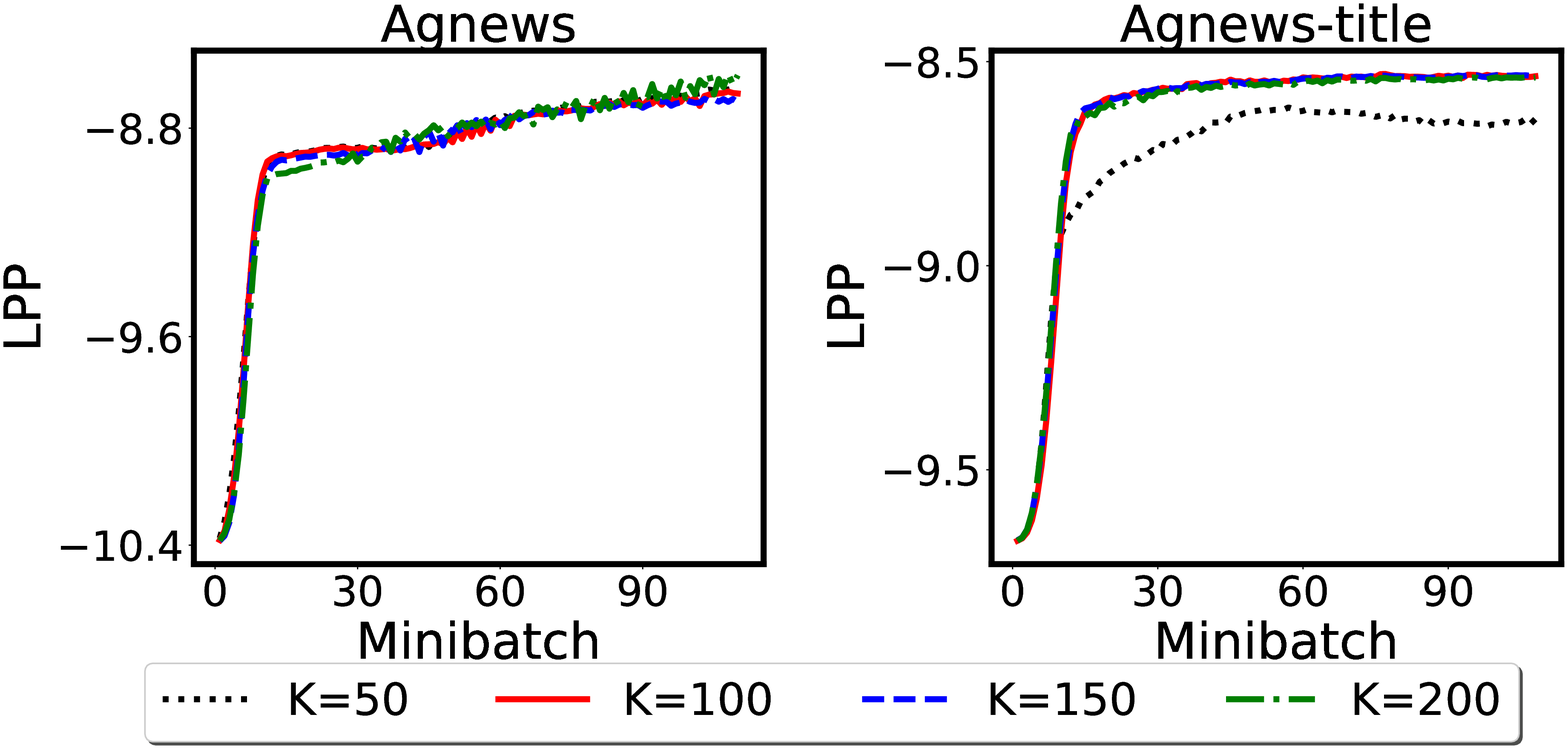} 
\end{center}
\caption{Sensitivity of GCTM-WN w.r.t the number of topics}
\label{sens_topic}
\end{figure*}


In this subsection, we examine the sensitivity of GCTM w.r.t. $\sigma$ and the number of topics $K$.  We use the scenario with fixed batchsize ($1000$) to conduct experiments on two datasets: Agnews and Agnews-title. We measure the LPP of GCTM-WN when one of these parameters is changed and the other is fixed.

\textit{The sensitivity of GCTM-WN w.r.t. $\sigma$}:  Figure \ref{sens_sigma} illustrates the experimental results when $K$ is fixed to $100$ and $\sigma$ is varied. It is obvious that the different values of $\sigma$ only make GCTM-WN vary in terms of LPP. Moreover, the effect of $\sigma$ is different between the short and regular text datasets. GCTM-WN on the short texts is more sensitive than itself on the regular texts. However, $\sigma=0.1$ ($\sigma^2=0.01$) makes the performances of GCTM-WN on both Agnews and Agnews-title the worst. $\sigma$ provides a way to adjust the impact of the global variables ($\beta$ and $\tilde{W}$) from a minibatch to the next one. The smaller $\sigma$ is, the stricter the constraint of the variables between two consecutive minibatches becomes. Therefore, a small value of $\sigma$ ($\sigma=0.1$) causes GCTM-WN to badly learn new knowledge from the current minibatch. 

\textit{The sensitivity of GCTM-WN w.r.t. $K$}: Figure \ref{sens_topic} illustrates the experimental results when the number of topics $K$ is varied and $\sigma$ is fixed to $1$. The LPPs of GCTM-WN are stable on the Agnews dataset when $K$ is changed. On the Agnews-title dataset, only $K=50$ makes the performance of GCTM-WN decrease. The more information of word co-occurrence leads LDA to reduce ambiguous topics, therefore, GCTM-WN on the regular texts is less sensitive than itself on the short texts. Moreover, when the short dataset is big, the number of topics $K$ should be large enough to achieve better performances.  

\section{Conclusion}

In conclusion, this paper proposes a novel model which integrates graph convolutional networks into a topic model to exploit a knowledge graph well. Moreover, a novel learning method is presented to simultaneously train both the networks and the topic model in streaming environments. It is worth noting that our method can be extended  for a wide class of probabilistic models. The extensive experiments show that our method can work well when dealing with short texts and concept drift. Our method significantly outperforms the state-of-the-art baselines in terms of generalization ability and topic coherence.


\bibliographystyle{cas-model2-names}
\bibliography{GCTM}

\appendix{}

\section{Log Predictive Probability}

We calculate log predictive probability on a test set as in \cite{lpp}. Let $D_{train}$ and $D_{test}$ be training and test sets respectively.  The model parameter $\beta$ of LDA is learnt on $D_{train}$. Each document in the test set $D_{test}$ is divided randomly into two disjoint parts $\mathbf{w_{obs}}$ and $\mathbf{w_{ho}}$ with a ratio of 80:20. LPP examine how a model predicts the words $\mathbf{w_{ho}}$ when giving the words $\mathbf{w_{obs}}$ for every document in the test set. The predictive probability is calculated as below:
\begin{align}
    p(\mathbf{w_{ho}} \mid \mathbf{w_{obs}}, \beta) &= \prod_{w \in \mathbf{w_{ho}}} p(w \mid \mathbf{w_{obs}}, \beta) \nonumber \\
    &\approx \prod_{w \in \mathbf{w_{ho}}} p(w \mid \theta^{obs}, \beta) \nonumber \\
    &= \prod_{w \in \mathbf{w_{ho}}} \sum_{k=1}^K p(w \mid z=k, \beta)p(z=k\mid \theta^{obs}) \nonumber \\
    &= \prod_{w \in \mathbf{w_{ho}}} \sum_{k=1}^K {\theta^{obs}_k\beta_{kw}} \nonumber
\end{align}
where $\theta^{obs}$ is inferred from $\mathbf{w_{obs}}$ and the learnt model $\beta$.
Then LPP of each document $d$ is computed:
\begin{align}
    LPP_d = \frac{\log p(\mathbf{w_{ho}} \mid \mathbf{w_{obs}}, \beta)}{|\mathbf{w_{ho}}|}
\end{align}
where $|\mathbf{w_{ho}}|$ is the length of $d$ in $\mathbf{w_{ho}}$). Then, the LPP of $D_{test}$ is averaged on all documents in the test set. We also run $5$ times with $5$ random splits to average. 

\section{Normalized Pointwise Mutual Information}

This metric was computed as in \cite{lau2014machine}. After training LDA, we pick top $t=20$ words with the highest probabilities in topic distribution ($\mathbf{w^k} = \{w^k_1, w^k_2,..., w^k_t\}$) for each topic $k$. We calculate NPMI of a topic $k$ as follows:
\begin{align}
    &\text{NPMI}(k, \mathbf{w^k}) = \frac{2}{t(t-1)} \sum_{i=2}^t\sum_{j=1}^{i-1}\frac{\log \frac{p(w^k_i, w^k_j)}{p(w^k_i)p(w^k_j)}}{-\log p(w^k_i, w^k_j)} \nonumber \\
    &\approx \frac{2}{t(t-1)} \sum_{i=2}^t\sum_{j=1}^{i-1}\frac{\log \frac{D(w^k_i, w^k_j) + 10^{-2}}{D} - \log \frac{D(w^k_i)D(w^k_j)}{D^2}}{-\log \frac{D(w^k_i, w^k_j) + 10^{-2}}{D}} \nonumber \\
    &= \frac{2}{t(t-1)} \sum_{i=2}^t\sum_{j=1}^{i-1} -1 + \frac{2\log D - \log D(w^k_i) - \log D(w^k_j)}{\log D - \log(D(w^k_i, w^k_j) + 10^{-2})} \nonumber
\end{align}
where $D$ is the total number of documents, $D(w^k_i)$ is the number of documents that contain $w^k_i$, $D(w^k_i, w^k_j)$ is the number of documents that contain both $w^k_i$ and  $w^k_j)$. Finally, NPMI is averaged on all $K$ topics.

\section{The effective settings of the methods}

In this section, we list the best hyperparameter for the methods from grid search.

\subsection{Experiments on datasets in terms of fixed batchsize, timestamp, and noisy data}

\textbf{PVB}: 

Yahoo-title: $\kappa= 0.9$, $S= 10^6$

NYtimes-title: $\kappa = 0.9$, $S =10^5$

Agnews: $\kappa = 0.9$, $S=10^4$

Agnews-title: $\kappa = 0.9$, $S=10^6$ 

TMN:  $\kappa = 0.9$, $S=10^3$

TMN-title:  $\kappa = 0.9$, $S=10^3$

Irishtimes (with timestamp): $\kappa = 0.5$, $S=10^5$

Irishtimes (with fixed batchsize): $\kappa = 0.9$, $S=10^5$

Twitter: $\kappa = 0.9$, $S=10^6$

\textbf{SVB-PP}: 

Yahoo-title: $\rho =0.99$

NYtimes-title: $\rho =0.99$

Agnews: $\rho =0.99$

Agnews-title: $\rho =0.99$

TMN: $\rho =0.99$

TMN-title: $\rho =0.99$

Irishtimes (with timestamp): $\rho =0.5$

Irishtimes (with fixed batchsize): $\rho =0.9$

Twitter:  $\rho =0.99$

\textbf{GCTM-WN}:

Yahoo-title: $\sigma =0.01$

NYtimes-title: $\sigma =100.0$

Agnews: $\sigma =1.0$

Agnews-title: $\sigma =1.0$

TMN: $\sigma =1.0$

TMN-title: $\sigma =1.0$

Irishtimes (with fixed batchsize): $\sigma = 0.01$

Irishtimes (with fixed batchsize): $\sigma =0.01$

Twitter: $\sigma =1$

\textbf{GCTM-W2V}:

Yahoo-title: $\sigma =100.0$

NYtimes-title: $\sigma =1.0$

Agnews: $\sigma =1.0$

Agnews-title: $\sigma =1.0$

TMN: $\sigma =100.0$

TMN-title: $\sigma =100.0$

Irishtimes (with fixed batchsize): $\sigma = 0.01$

Irishtimes (with fixed batchsize): $\sigma =0.01$

Twitter: $\sigma =1$
\subsection{Experiments on datasets in terms of concept drift and catastrophic forgetting}

PVB:  $\kappa = 0.9$, $S=10^6$

SVB-PP: $\rho =0.9$

GCTM-WN: $\sigma =100$

GCTM-W2V: $\sigma =0.01$

\end{document}